\def\year{2019}\relax
\documentclass[letterpaper]{article} 
\usepackage{aaai19}  
\usepackage{times}  
\usepackage{helvet}  
\usepackage{courier}  
\usepackage{url}  
\usepackage{graphicx}  
\title{From Independent Prediction to Reordered Prediction: Integrating Relative Position and Global Label Information to Emotion Cause Identification}

\author{
	Zixiang Ding, Huihui He, Mengran Zhang, Rui Xia\thanks{The corresponding author of this paper.}\\
	School of Computer Science and Engineering, Nanjing University of Science and Technology, China\\
	dingzixiang@njust.edu.cn, hehuihui1994@gmail.com, zhangmengran@njust.edu.cn, rxia@njust.edu.cn
}

\begin{document}
\maketitle
\begin{abstract}
 Emotion cause identification aims at identifying the potential causes that lead to a certain emotion expression in text. Several techniques including rule based methods and traditional machine learning methods have been proposed to address this problem based on manually designed rules and features. More recently, some deep learning methods have also been applied to this task, with the attempt to automatically capture the causal relationship of emotion and its causes embodied in the text. In this work, we find that in addition to the content of the text, there are another two kinds of information, namely relative position and global labels, that are also very important for emotion cause identification. To integrate such information, we propose a model based on the neural network architecture to encode the three elements ($i.e.$, text content, relative position and global label), in an unified and end-to-end fashion. We introduce a relative position augmented embedding learning algorithm, and transform the task from an independent prediction problem to a reordered prediction problem, where the dynamic global label information is incorporated. Experimental results on a benchmark emotion cause dataset show that our model achieves new state-of-the-art performance and performs significantly better than a number of competitive baselines. Further analysis shows the effectiveness of the relative position augmented embedding learning algorithm and the reordered prediction mechanism with dynamic global labels.
\end{abstract}

\section{Introduction}
Emotion cause identification, as a sub-task of emotion analysis, aims at identifying the potential causes that lead to a certain emotion expression in text. It has gained much attention in recent years due to its wide potential applications \cite{lee2010text,russo2011emocause,gao2015emotion,cheng2017emotion}. In the benchmark emotion cause dataset released by \cite{gui2016event}, emotion cause identification was formally defined as a binary clause classification problem that detects for each clause in the document whether this clause contains potential emotion causes given the annotation of emotion. In this paper, we refer to a instance of the corpus as a ``document", although it is normally a short document.

For example, in the following document:

\noindent \textit{Yesterday morning, a policeman visited the old man with the lost money, and told him that $<$cause$>$ the thief was caught $<$/cause$>$. The old man was very $<$emotion word$>$ happy $<$/emotion word$>$. Accompanied by the policeman, he deposited the money in the bank.}

\noindent  There are six clauses in this document. The emotion ``happy" is contained in the fourth clause (we denote this clause as ``emotion clause", which refers to a clause that contains emotion expressions), and the corresponding cause ``the thief was caught" is in the third clause. The emotion cause identification task was actually defined as a clause-level binary classification problem in \cite{gui2016event,gui2017question}. The goal is to detect for each clause in a document, whether the clause contains potential emotion causes ($i.e.$, whether the clause is an emotion cause) given the annotation of emotion. As \cite{gui2016event} pointed, this task is more difficult than emotion classification because it requires a deeper understanding of the document.

In the past, several techniques including rule based methods and traditional machine learning methods have been proposed to address this problem based on manually designed rules and features \cite{lee2010text,lee2013detecting,chen2010emotion,gui2016event}. More recently, some deep neural networks have also been applied to this task, with the attempt to automatically capture the causal relationship of emotion and its causes. For example, \cite{cheng2017emotion} proposed to use a long short-term memory (LSTM) network to address this task. \cite{gui2017question} introduced a new deep memory network architecture which encodes the context of each word by multiple memory slots.

While most of these methods emphasized on the usage of content information, we find in this work, in addition to the content information, there are two other kinds of information that are also very important for emotion cause identification:
\begin{itemize}
	\item ``relative position", which denotes the relative distance to the emotion clause;
	\item ``global label", which means the predicted labels of the other clauses in the whole document.
\end{itemize}

However, most of the previous work including the state-of-the-art predicted clauses independently and ignored the global label information. It was hence possible that none of the clauses in the document was predicted as the emotion cause, or too many clauses in the same document were predicted as emotion causes. Although some of them have already used the position information in their neural networks for emotion cause identification, but their methods were relatively simple. For example, \cite{gui2016event,gui2017question,xu2017ensemble} concatenated position embedding with the representation of a clause. To the best of our knowledge, none of the previous work has explicitly used the global label information for emotion cause identification.

In this paper, we propose a unified neural architecture for emotion cause identification that takes the relative position and global label information into account. Firstly, we attach a relative position embedding to word embedding, and propose an encode-decode-style module to help learn a relative position augmented embedding. Secondly, we sort the clauses according to the absolute value of relative position in an ascending order, and transform the emotion cause identification task from an independent prediction problem into a reordered prediction problem. The predictions of the previous clauses are used as features for predicting the subsequent clauses in a dynamic manner, so as to integrate the global label information.

We evaluated our approach on the benchmark emotion cause dataset \cite{gui2016event}. The experimental results indicate that our model achieves new state-of-the-art performance in emotion cause identification and performs significantly better than a number of competitive baselines. Further in-depth analysis has proved the effectiveness of the relative position augmented embedding learning algorithm and the reordered prediction mechanism with dynamic global labels.

\section{Related Work}
\cite{lee2010text} first presented a task on emotion cause identification. They manually constructed a small-scale corpus from the Academia Sinica Balanced Chinese Corpus. Based on this corpus, \cite{lee2013detecting} developed a rule-based approach to detect emotion causes. \cite{neviarouskaya2013extracting} also explored linguistic relations of emotion cause by applying syntactic and dependency parser. Some studies extended rule-based approaches to informal texts such as microblog texts. \cite{li2014text} tried to extract the reasons of emotions by introducing knowledge and theories from other fields such as sociology. Based on this idea, they built an automatic rule-based system to detect the cause event of each emotional microblog post. \cite{gao2015emotion} and \cite{gao2015rule} designed a set of complex rules considering a cognitive emotion model and emotions categories to extract emotion cause in Chinese microblogs. More recently, \cite{yada2017bootstrap} proposed a bootstrapping technique to automatically obtain conjunctive phrases as textual cue patterns for emotion cause identification.

Most of the above work is based on rules, there were also some machine learning based methods. \cite{chen2010emotion} and \cite{gui2014emotion} used the extended linguistic rules of \cite{lee2010text} as features, and chose machine learning model, such as SVM, to detect emotion causes. \cite{russo2011emocause} proposed a crowd-sourcing approach to build a common-sense knowledge base which is related to emotion causes. However, automatically expanding the common sense knowledge base is very challenging. \cite{ghazi2015detecting} used conditional random fields (CRFs) to label the semantic roles on the emotion-related text from an English corpus. \cite{song2015detecting}  utilized context-sensitive topical pagerank to detect meaningful multi-word expressions as emotion causes. Recently, some researchers explored structure-based representation of events to improve the performance of emotion cause detection. \cite{gui2016event} and \cite{gui2016emotion} released an emotion cause dataset and proposed a seven-tuple representation structure to describe emotion cause events and employed SVM-based methods. \cite{xu2017ensemble} proposed an emotion cause detection method by using event extraction framework, in which a tree structure-based representation method is used to represent the events.

Deep learning techniques have also been applied to emotion cause identification. \cite{cheng2017emotion} used long short-term memory (LSTM) for emotion cause detection. \cite{gui2017question} proposed a new deep memory network architecture to model the context of each word simultaneously by multiple memory slots. In this work, we also address the emotion cause identification task under deep neural networks.  We emphasize on the usage of relative position and global label information for emotion cause identification. Although some existing work have also used the position information. However, they simply used the position as a feature or embedding for training their model \cite{gui2016event,gui2017question,xu2017ensemble}. The global label information were also ignored in their methods. In comparison, we use relative position and global label in a novel and more efficient manner: 1) We propose an encode-decode-style method to incorporate relative position information into clause representation. 2) We predict the clauses in the same document in a special order according to the relative position, and update ``dynamic global label" with previous predictions.

\section{Motivation}
The motivation of this paper is based on the following two observations.

In Table 1, we first summarize the percentage of a clause being emotion cause at different positions, based on the emotion cause benchmark corpus \cite{gui2016event}. We label the position of the emotion clause as $0$, and use ``relative position" to denotes the relative distance to the emotion clause. For example,  $-1$ denotes one clause left to the emotion clause; $+2$ denotes two clauses right to the emotion clause. It can be observed that the clauses that are closer to the emotion clause are more likely to be an emotion cause. In detail, the clause at position $-1$ has the highest probability of being an emotion cause. The probabilities of clauses with larger distance reduce gradually. The ascending order of the distance is $0, -1, +1, -2, +2, -3, +3, \cdots$. This is natural since people are accustomed to describe the cause near the emotion expression.

\begin{table}
	
	\centering
	\caption{\label{font-table} The percentage of a clause being emotion cause at different relative positions. }
	\begin{tabular} {c|c|c}
		\hline  Relative Position & Number & Percentage\\
		\hline
		-3 & 37 & 1.71\% \\
		-2 &	167 &	7.71\% \\
		-1 &	1,180 &	54.45\% \\
		0 &	511 &	23.58\% \\
		+1 &	162 &	7.47\% \\
		+2	& 48 &	2.22\% \\
		+3	& 11 &	0.51\% \\
		Others &	42 &	1.94\% \\
		\hline
	\end{tabular}
	\label{tab:table1}
\end{table}

\begin{table}
	
	\centering
	\caption{\label{font-table} The proportion of documents with different number of emotion causes. }
	\begin{tabular} {c|c|c}
		\hline   & Number & Percentage\\
		\hline
		Document with one cause & 2046 & 97.20\% \\
		Document with two causes &	56 &	2.66\% \\
		Document with three causes &	3 &	0.14\% \\
		All &	2105 &	100\% \\
		\hline
	\end{tabular}
	\label{tab:table1}
\end{table}

In Table 2, we furthermore summarize the proportion of documents with different number of emotion causes. In each document, only one emotion was annotated. It can be seen that 97.20\% of the documents have only one corresponding emotion cause. Only a small number of documents have two or three corresponding emotion causes. The percentages are 2.66\% and 0.14\% respectively. Documents with more than three emotion causes do not appear in the corpus. However, most of the previous work predicted clauses independently and ignore the global label information. It is hence possible that none of the clauses in the document is predicted as an emotion cause, or too many clauses in the same document are predicted as emotion causes.

Jointly observing Table 1 and Table 2, we infer that in addition to relative position, the global label information is also very important for predicting individual clauses. Let us consider the ascending order of relative positions according to their absolute value ($i.e.$, $0, -1, +1, -2, +2, -3, +3, \cdots$). If the clause in the front (which is closer to the emotion clause) was predicted to be a cause with a high probability, the probability of subsequent clause being a cause should be reduced, as most of the documents have only one cause. On the contrary, if none of the previous clause was predicted as an emotion cause, the probability of subsequent clause should be increased.

In summary, we conclude based on the above analysis that
\begin{itemize}
	\item the relative position information ($i.e.$, the relative distance to the emotion clause) plays an important role in emotion cause identification;
	\item the global label information ($i.e.$, the predictions of other clauses) must be considered when predicting individual clause.
\end{itemize}

\begin{figure*}[!t]
	\centering
	\includegraphics[width=2\columnwidth ]{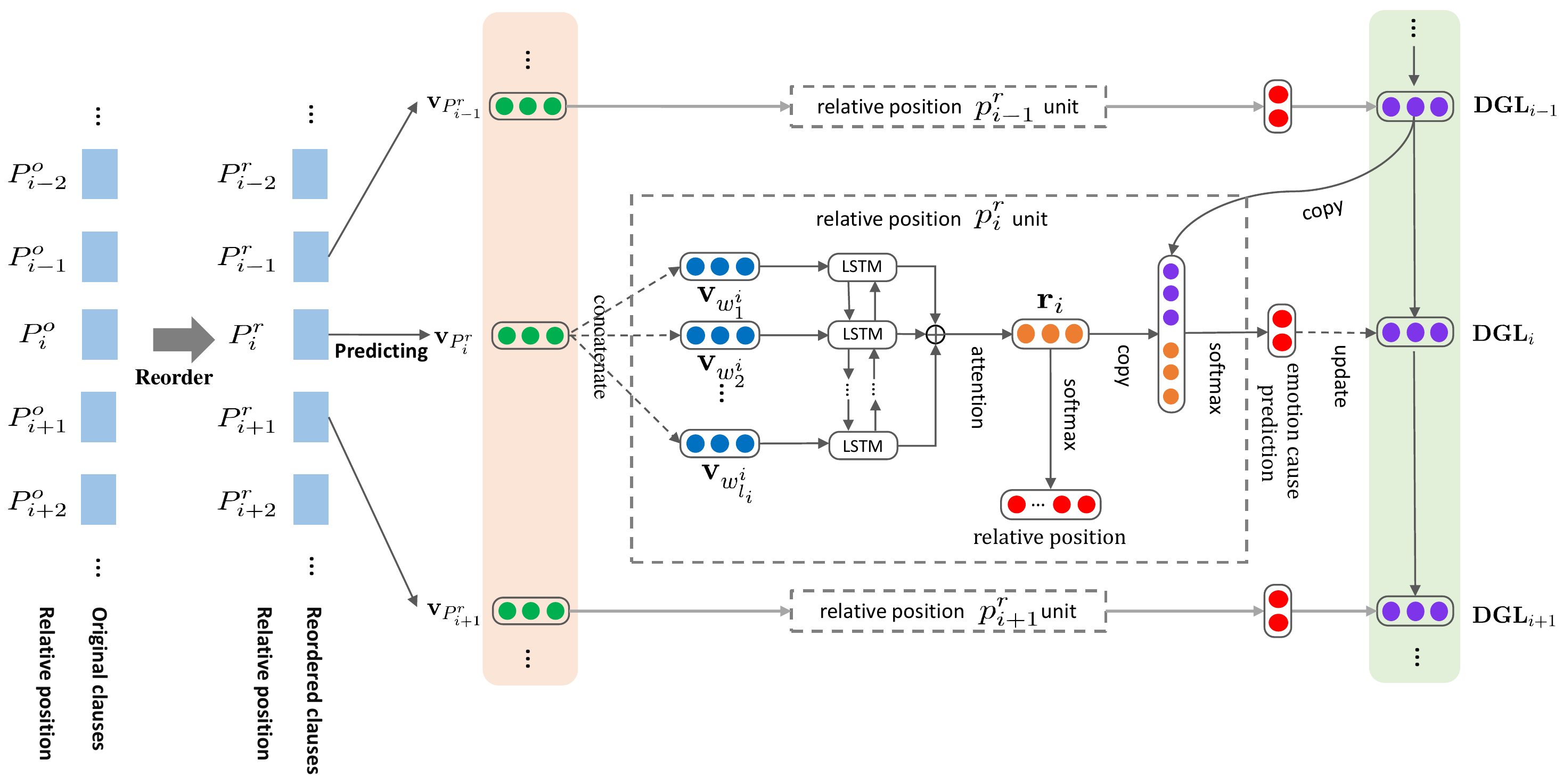}
	\caption{\label{font-table}Overview of PAE-DGL model.}
	\label{fig_sim1}
\end{figure*}

\section{Model}
In this section, we first introduce the overall architecture of our model. Then we describe the ``relative position augmented embedding learning" (PAE for short) module and the ``reordered prediction with dynamic global labels" (DGL for short) module in detail.

\subsection{Overall Description}

The overall architecture of the proposed method is shown in Figure 1. Following the state-of-the-art emotion cause identification works \cite{gui2017question}, we consider the task as a clause-level binary classification problem. For each clause in a document, the model predicts whether this clause is the emotion cause.

Formally, a document consisting of $N$ clauses can be represented as $D = \{c_1,c_2,\cdots,c_N\} $ , where $c_i$ is the $i$th clause. The $i$th clause consisting of $l_i$  words can be represented as $c_i = \{w^i_1,w^i_2,\cdots,w^i_{l_i}\} $. Each word $w^i_j$ is represented as a word embedding $ \mathbf{v}_{w^i_j} \in R^{m} $. In order to capture the text content information, we employ a Bi-directional Long Short-Term Memory (Bi-LSTM) structure to encode each clause separately. Specifically, we feed $c_i$ into Bi-LSTM and get hidden states $\{\mathbf{h}^i_1,\mathbf{h}^i_2,\cdots,\mathbf{h}^i_{l_i}\} $, where $ \mathbf{h}^i_j \in R^d$ is the hidden state of the $j$th word in $c_i$. Then we adopt the attention mechanism to get the clause representation $\mathbf{r}_i$, where $ \mathbf{r}_i \in R^d$ is a weighted sum of hidden states. Here we omit the details of Bi-LSTM and attention for limited space, readers can refer to \cite{graves2013speech} and \cite{bahdanau2014neural}.

The above standard Bi-LSTM and attention architecture only encode the content information. To incorporate the relative position and global label information, we introduce two mechanisms into the network.

Firstly, we attach a relative position embedding to word embedding, and proposed an encode-decode-style module by using relative position as supervision to help learn a relative position augmented embedding.

Secondly, we sort the clauses according to the absolute value of relative position in an ascending order, and transform the task from an independent prediction problem into a reordered prediction problem. The predictions of the previous clauses are used as features for predicting the subsequent clauses in a dynamic manner.

\subsection{Relative Position Augmented Embedding Learning}

We define the relative position of the  $i$th clause as  $P_i$, where  $P_i \in \{ \cdots,-3,-2,-1,0, +1, +2, +3,\cdots\} $. Then we propose to learn an embedding vector for each relative position.

Firstly, we attach the position embedding to each word embedding to shape a position-augmented word embedding. We use $ \mathbf{v}_{P_i} \in R^{n} $ to represent the embedding of relative position $P_i$, where $n$ is the dimension of the embedding. The position-augmented embedding of a word $ w^i_j $ for a specific relative position $P_i$ is:
\begin{equation}
\label{eqn_example}
\mathbf{e}_{w^i_j} = \mathbf{v}_{w^i_j} \oplus  \mathbf{v}_{P_i},
\end{equation}
where operator $ \oplus $ indicates the concatenation operation, thus the dimension of $ \mathbf{e}_{w^i_j}  $ is $ (m+n)  $. Then the position-augmented word embedding of each word in $c_i$ is feed into Bi-LSTM and attention module to get the position-augmented clause representation $\mathbf{r}_i$.

Secondly, we introduce a relative position embedding learning module by using relative position as a kind of supervision, to ensure that the relative position information can be better encoded. Specifically, we take the clause representation $\mathbf{r}_i$ as feature to predict the relative position of clauses:
\begin{equation}
\label{eqn_example}
\mathbf{\hat{p}}_i = \textrm{softmax} ( \mathbf{W}_s \mathbf{r}_i+\mathbf{b}_s ),
\end{equation}
where $ \mathbf{W}_s \in R^{|D| \times d} $ and $ \mathbf{b}_s \in R^{|D|}$ are weight matrix and bias for softmax classification. $ \mathbf{\hat{p}}_i \in R^{|D|}$ is the predicted distribution of relative position, where $ |D| $ is the number of different relative positions.

The cross-entropy error of relative position prediction on a document is defined as $ loss^p $:
\begin{equation}
\label{eqn_example}
loss^p = -\sum_{i=1}^{N} \sum_{j=0}^{|D|-1} p_{i,j} \cdot \textrm{log}(\hat{p}_{i,j}),
\end{equation}
where $N$ denotes the number of clauses in the document, and $ \mathbf{p}_i  $ is the ground truth distribution of relative position.

The learning algorithm can be actually regarded as a generalized autoencoder. The autoencoder consists of encoder and decoder, and learn the codings in an unsupervised manner. Our method does not uncompress the middle code into the original signal (relative position embeddings) during decoding, but learns the codings under the supervision of relative position value.

\begin{figure*}[!t]
	
	\centering
	\includegraphics[width=\textwidth ]{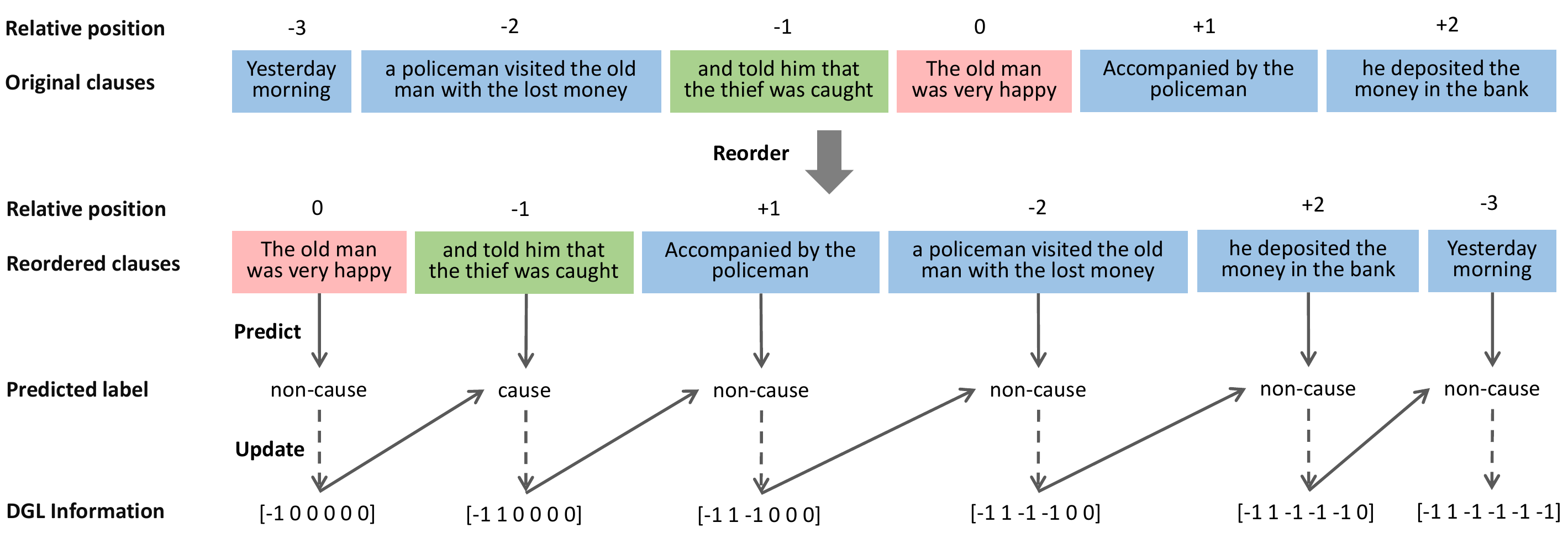}
	\caption{\label{font-table} An example of clause reordering with dynamic global labels.}
	\label{fig_sim2}
\end{figure*}

\subsection{Reordered Prediction with Dynamic Global Labels}

As we have mentioned, most of the previous work predicted clauses independently and ignored global label information. It is hence possible that none of the clauses in the document is predicted as emotion causes, or too many clauses in the same document are predicted as emotion causes. To address this, we propose to sort the clauses according to relative position first and then integrate the global label information in a dynamic manner.

\subsubsection{Reordered Prediction}

Firstly, we sort the clauses according to the absolute value of relative position in an ascending order ($i.e., 0, -1, +1, -2, +2, -3, +3, \cdots$). The order of clauses before and after reordering is denoted as $ \mathbf{P^o} $ and $ \mathbf{P^r} $ respectively. Taking Figure 2 as an example, $ \mathbf{P^o} $ is : $[-3,-2, -1, 0, +1, +2]$, after reordering, $ \mathbf{P^r} $ is : $[0, -1, +1, -2, +2, -3]$.

We then transform the emotion cause identification task from an independent prediction problem into a reordered prediction problem. The clauses in a document are predicted sequentially according to the order of $ \mathbf{P^r} $. In this order, the clauses which are closer to the emotion clause (more likely to be an emotion cause) are predicted first, and the clauses which have a larger distance to the emotion clause (less likely to be an emotion cause) are predicted later.

\subsubsection{Dynamic Global Label}

In the order of $ \mathbf{P^r} $, if a clause in the front was predicted to be emotion cause with a high probability, the probability the subsequent clause being an emotion cause should be reduced, because most of the documents have only one emotion cause (according to Table 1). On the contrary, if none of the previous clause was predicted as an emotion cause, the probability the subsequent clause should be increased.

To meet this end, we propose a dynamic global label feature template to record the predictions of the previous clauses to help the prediction of the subsequent clause.

Formally, we use a vector $\mathbf{DGL} \in R^{q} $ to represent the prediction of all clauses (reordered) in a document, where $q$ serves as the maximum length of documents.

Firstly, the vector $\mathbf{DGL}$ dynamically changes during the reordered prediction of clauses. For a document consisting of $N$ clauses, our model predict for $N$ times. For the first prediction, the vector $\mathbf{DGL}$ is set to $\mathbf{DGL}_0=[0,0,0,\cdots]$, then the clause at relative position $ P^r_0 $ is predicted. If the clause is predicted as emotion cause, then $\mathbf{DGL}$ is set to $\mathbf{DGL}_1=[1,0,0,\cdots]$, otherwise $\mathbf{DGL}_1=[-1,0,0,\cdots]  $. Then we predict the clause at relative position $ P^r_1 $, and update $\mathbf{DGL}$ to $\mathbf{DGL}_2 $ according to the prediction result of current clause. We repeat this process until all clauses are predicted.

Secondly, for the prediction of $i$th clause $c_i$, we take the concatenation of $\mathbf{r}_i$ and $\mathbf{DGL}_{i-1}  $ as features for classification. $\mathbf{r}_i$ encodes the text content and relative position information, and $\mathbf{DGL}_{i-1}  $ encodes the global label information. The prediction of the clause is calculated as follows:
\begin{equation}
\label{eqn_example}
\mathbf{\hat{y}}_i = \textrm{softmax} ( \mathbf{W}_c [\mathbf{r}_i;\mathbf{DGL}_{i-1}]+\mathbf{b}_c ),
\end{equation}
where $ [\mathbf{r}_i;\mathbf{DGL}_{i-1}] \in R^{d+q}$ is concatenation of the two vectors. $ \mathbf{W}_c \in R^{2 \times (d+q)} $ and $ \mathbf{b}_c \in R^{2}$ are weight matrix and bias. $ \mathbf{\hat{y}}_i \in R^{2}$ is the predicted distribution of emotion cause.

The cross-entropy error of emotion cause prediction on the whole document is defined as $ loss^c $:
\begin{equation}
\label{eqn_example}
loss^c = -\sum_{i=1}^{N} \sum_{j=0}^{1} y_{i,j} \cdot \textrm{log}(\hat{y}_{i,j}),
\end{equation}
where $ \mathbf{y}_i  $ is the ground truth distribution of emotion cause of the $i$th clause.

The final loss of our model is a weighted sum of $ loss^p  $ and $ loss^c  $ with L2-regularization term as follows:
\begin{equation}
\label{eqn_example}
loss = \lambda_p loss^p + \lambda_c loss^c + \lambda ||\theta||^2,
\end{equation}
where $ \lambda_p $, $ \lambda_c $, $ \lambda $ is the weight of $ loss^p  $, $ loss^c  $, L2-regularization term respectively and $ \theta $ denotes the parameter set.
It should be noted that the dynamic global label is updated by true labels of each clause in the training phase to accelerate the training process.

\section{Experiments}
\subsection{Dataset and Experimental Settings}

We evaluated our proposed model on a Chinese emotion cause corpus \cite{gui2016event} \footnote{http://hlt.hitsz.edu.cn/?page\_id=694}, which is the mostly used dataset for emotion cause identification. The same as \cite{gui2017question}, we stochastically select 90\% of the data as training data and the remaining 10\% as testing data. Also, in order to obtain statistically credible results, we repeat the experiments 25 times and report the average result.
The precision, recall, and F score are used as the metrics for evaluation. These metrics in emotion cause identification are defined by:
\begin{equation}
\label{eqn_example}
P= \frac {\sum correct\_causes} {\sum proposed\_causes},
\end{equation}
\begin{equation}
\label{eqn_example}
R= \frac {\sum correct\_causes} {\sum annotated\_causes},
\end{equation}
\begin{equation}
\label{eqn_example}
F= \frac {2 \times P \times R} { P + R} .
\end{equation}
$proposed\_causes$ denotes the number of clauses that are predicted to be emotion cause, $annotated\_causes$ denotes the number of clauses that are labeled as emotion cause and the $correct\_causes$ means the number of clause that are both labeled as emotion cause and predicted to be emotion cause.

We use word vectors that were pre-trained on the 1.1 million Chinese Weibo corpora provided by NLPCC2017 \footnote{http://www.aihuang.org/p/challenge.html} with word2vec\footnote{ https://code.google.com/archive/p/word2vec/} \cite{mikolov2013distributed} toolkit. Similar performance can be obtained by using the embedding provided in \cite{gui2017question}. The dimensions of word embedding and position embedding are set to 200 and 50, respectively. The hidden units of LSTM is set to 100. All weight matrices and bias are randomly initialized by a uniform distribution $U(-0.01, 0.01)$. The code has been made publicly available through Github \footnote{https://github.com/NUSTM/PAEDGL}.

\begin{table}
	
	\centering
	\caption{\label{font-table} Comparison with existing methods using precision, recall, and F-measure as metrics. The best result of each column is highlighted in bold. }
	\begin{tabular} {c|c|c|c}
		\hline    & P & R & F\\
		\hline
		RB & 0.6747 & 0.4287 & 0.5243 \\
		CB & 0.2672 & \textbf{0.7130} & 0.3887 \\
		RB+CB & 0.5435 & 0.5307 & 0.5370 \\
		RB+CB+ML & 0.5921 & 0.5307 & 0.5597 \\
		SVM & 0.4200 & 0.4375 & 0.4285 \\
		Word2vec & 0.4301 & 0.4233 & 0.4136 \\
		CNN & 0.6215 & 0.5944 & 0.6076 \\
		Multi-Kernel & 0.6588 & 0.6927 & 0.6752 \\
		Memnet & 0.5922 & 0.6354 & 0.6131 \\
		ConvMS-Memnet & 0.7076 & 0.6838 & 0.6955 \\
		\hline
		PAE-DGL & \textbf{0.7619} & 0.6908 & \textbf{0.7242} \\
		\hline
	\end{tabular}
	\label{tab:table1}
\end{table}

\subsection{Evaluation and Comparison}

We compare our model with the following baseline methods:

\begin{itemize}
	\item \textbf{RB} (Rule based method): The method based on manually defined linguistic rules proposed in \cite{lee2010text};
	\item \textbf{CB} (Common-sense based method): The approach is based on knowledge and is proposed by \cite{russo2011emocause};
	\item \textbf{RB+CB+ML} (Machine learning method trained from rule-based features and common-sense knowledge base): This method uses rules and facts in a knowledge base as features for classification, which is proposed by \cite{chen2010emotion};
	\item \textbf{SVM}: This is a SVM classifier that uses the unigram, bigram and trigram features. It is a baseline previously used in \cite{gui2017question};
	\item \textbf{Word2vec}: This is a SVM classifier using word representations learned by Word2vec as features;
	\item \textbf{Multi-kernel}: This method uses the multi-kernel method \cite{gui2016event} to identify the emotion cause;
	\item \textbf{CNN}: The convolutional neural network for sentence classification \cite{kim2014convolutional};
	\item \textbf{Memnet}: The deep memory network proposes by \cite{gui2017question}. Word embeddings are pre-trained by skip-grams and the number of hops is set to 3;
	\item \textbf{ConvMS-Memnet}: The convolutional multiple-slot deep memory network proposes by \cite{gui2017question}. It is the state-of-the-art method for emotion cause identification. We use the best performance reported in their paper.
\end{itemize}

Results of PAE-DGL and baseline methods are listed in Table 3. It can be seen that our method achieves significant improvement in precision, which is 5.43\% higher than the best result of previous methods. As for the results of recall, CB (common-sense based method) performs the best and reaches 71.3\%. However, its precision is quite low, resulting in poor performance in F-measure. Finally, our method outperforms the state-of-the-art method ConvMS-Memnet by 2.87\% in F-measure. This shows that our model can better identify emotion causes by using text content, relative position and global label information reasonably.

\subsection{The Effect of PAE and DGL}

To further explore the factors that bring improvement to the model, we designed three experiments:

\begin{itemize}
	\item \textbf{Bi-LSTM}: We only use Bi-LSTM with attention to get clause representation, which is the only feature we use for classification. This method only uses content information
	\item \textbf{PAE}: We add PAE module to the Bi-LSTM model, and denote this method as PAE. This method uses text content and relative position information.
	\item \textbf{PAE-DGL}: This is the complete model proposed in this paper, which jointly uses the text content, relative position and global label information.
\end{itemize}

\begin{table}
	
	\centering
	\caption{\label{font-table} The effect of PAE and DGL. }
	\begin{tabular} {c|c|c|c}
		\hline    & P & R & F\\
		\hline
		Bi-LSTM & 0.5445 & 0.1663 & 0.2529 \\
		PAE & 0.6897 & 0.6794 & 0.6836 \\
		PAE-DGL & \textbf{0.7619} & \textbf{0.6908} & \textbf{0.7242} \\
		\hline
	\end{tabular}
	\label{tab:table1}
\end{table}

The results are listed in Table 4. We can obviously find that Bi-LSTM performs the worst, which indicates the text content information is not enough for emotion cause identification. PAE further incorporates position information based on Bi-LSTM and obtain significant improvement in F-measure. Moreover, based on PAE, we further incorporates dynamic label information in PAE-DGL, and the F-measure improves by 4.06\%. In addition, we conducted an in-depth analysis on what is gained by using the relative position information and global label.

By observing the documents that are correctly classified by PAE but misclassified (contain misclassified clauses) by BiLSTM, we found that most of these documents were predicted to be without any cause by BiLSTM. Further quantitative analysis revealed that approximately 65.4\% of documents were predicted by BiLSTM as not containing any cause. In contrast, this ratio drops to 12.8\% for PAE. To conclude, the relative position plays an indicative role for predicting emotion causes, which can significantly improve the recall.

By observing the documents that are correctly classified by PAE-DGL but misclassified by PAE, we found that most of these documents were predicted by PAE as containing two causes, however, only one of the two causes was correctly predicted. To conclude, the global label information can effectively avoid predicting too many causes in the same document, which can significantly improve the precision.

\begin{table}
	
	\centering
	\caption{\label{font-table} Evaluation of different ways of modeling positions. }
	\begin{tabular} {c|c|c|c}
		\hline    & P & R & F\\
		\hline
		PL & 0.7018 & 0.6496 & 0.6743 \\
		PEC & \textbf{0.7081} & 0.5867 & 0.6405 \\
		PAE & 0.6897 & \textbf{0.6794} & \textbf{0.6836} \\
		\hline
	\end{tabular}
	\label{tab:table1}
\end{table}

\subsection{The Evaluation of Different Ways of Modeling Positions}

In this section, we explore different ways of using relative position information. The easiest way is to use the relative position as the last word of the clause, and learn a position embedding with the same dimension as word embedding. We call this way as position serving as the last word (PL). Another way is to take position embedding as feature for classification, that is, to concatenate the position embedding into the representation of the clause. We call this way as position embedding in clause level (PEC). PEC is the way used by \cite{gui2016event,gui2017question,xu2017ensemble}. The last way of using position information is our proposed PAE.

Table 5 shows the results of PL, PEC and PAE. Here we use the Bi-LSTM to encode each word in the clauses. Thus the dimension of the clause representation equals to the hidden units of Bi-LSTM. However, PEC concatenates the position embedding into the representation of the clause, thus the feature dimension is much bigger than that of the other methods and may cause over-fitting. In contrast, PAE incorporates position information without increasing feature dimensions. By taking the encode-decode-style, each clause representation is forced to encode more position information. According to Table 5, PAE performs better than the other methods, which indicates that PAE can make use of position information more sufficiently.

\subsection{Discussions on Reordered Prediction with DGL Information}

In order to explore the impact of using different ways of clause reordering on the model, we conduct the following experiments:

\begin{itemize}
	\item \textbf{DGL-$\mathbf{P}^o$}: We do not reorder the clauses in the PAE-DGL model. Instead, we make predictions based on the order that they appear in the document.
	\item \textbf{DGL-$\mathbf{P}^r$}: This is the complete model proposed in this paper. We predict the clauses which are closer to the emotion clause first.
\end{itemize}

\begin{table}
	
	\centering
	\caption{\label{font-table} Evaluation of different orders for DGL. }
	\begin{tabular} {c|c|c|c}
		\hline    & P & R & F\\
		\hline
		DGL-$\mathbf{P}^o$ & 0.6997 & 0.6561 & 0.6764 \\
		DGL-$\mathbf{P}^r$ & \textbf{0.7619} & \textbf{0.6908} & \textbf{0.7242} \\
		\hline
	\end{tabular}
	\label{tab:table1}
\end{table}

Table 6 shows the experimental results of PAE-DGL with different orders. PAE-DGL performs much better than PAE-DGL and achieves 4.78\% improvement in F-measure. This indicates that the order of clauses plays an important role on the task of emotion cause identification. It also shows that the proposed order is significantly better than the default order.

\begin{table}
	
	\centering
	\caption{\label{font-table} Different dynamic global labels in DGL. }
	\begin{tabular} {c|c|c|c}
		\hline    & P & R & F\\
		\hline
		DGL & \textbf{0.7412} & 0.6866 & 0.7129 \\
		DGL-Upper-Bound & 0.7402 & \textbf{0.7880} & \textbf{0.7633} \\
		\hline
	\end{tabular}
	\label{tab:table1}
\end{table}

\subsubsection{Exploring The Upper Bound of DGL}

In the above experiments, the DGL feature in the training phase is obtained by using true labels, and the DGL feature in the test phase is based on the predicted labels.

We further conduct an experiment by using the true labels of the test data to construct the DGL feature. Note that this can not be applied to real application. Our goal is only to explore the upper bound of the effect of DGL feature. Specifically, we comprare the following two models:

\begin{itemize}
	\item \textbf{DGL}: The DGL feature in the test phase is based on the predicted labels.
	\item \textbf{DGL-Upper-Bound}: The DGL feature in the test phase is obtained by using true labels.
\end{itemize}

The experimental results are shown in Table 7. We can find that the F-measure of DGL-Upper-Bound is 5.04\% higher than that of DGL. This further shows that 1) the prediction of the clauses in the same document can affect each other, and the DGL module can effectively capture this constraint information, and 2) knowing the overall prediction of the current document can effectively guide the prediction of the current clause.

\section{Conclusions}

In this work, we find that in addition to the content information, the relative position and global label information are also important factors for emotion cause identification. However, the use of the such information in previous work was either too simple or completely ignored. We propose a model based on the neural network architecture to incorporate three factors ($i.e.$, content, relative position and global label) in an unified and end-to-end fashion. The relation position is encoded as an embedding attached to the word embedding, an encode-decode-style module was further proposed to help better learn the embedding by using relative position as supervision. The clauses in a document is then sorted according to their relative positions. Instead of predicting clauses independently, we consider the emotion cause identification task as a reordered clause prediction problem. The predictions of the previous clauses are used as a kind of global label information for predicting the subsequent clauses in a dynamic manner. The experimental results on a benchmark emotion cause dataset showed that our approach achieves new state-of-the-art performance in emotion cause identification and performs significantly better than a number of competitive baselines.

\section{Acknowledgments}
\noindent Zixiang Ding and Huihui He contributed equally to this paper. Rui Xia was the main contributor to the idea of this paper. The work was supported by the Natural Science Foundation of China (No. 61672288), and the Natural Science Foundation of Jiangsu Province for Excellent Young Scholars (No. BK20160085).

\bibliographystyle{aaai}
\bibliography{dzx}

\end{document}